\documentclass[runningheads]{llncs}
\usepackage{graphicx}

\usepackage{tikz}
\usepackage{comment}
\usepackage{amsmath,amssymb} 
\usepackage{color}

\usepackage[accsupp]{axessibility}  

\usepackage{caption}
\usepackage{subcaption}
\usepackage{float}
\usepackage{hyperref}

\begin{document}
\setlength{\abovecaptionskip}{0.1em}
\setlength{\belowcaptionskip}{-1em}
\pagestyle{headings}
\mainmatter
\def\ECCVSubNumber{30}  

\title{Patient-level Microsatellite Stability Assessment from Whole Slide Images By Combining Momentum Contrast 
Learning and Group Patch Embeddings} 


\titlerunning{MS Status Classification using MoCo and Group Patch Emeddings}
%

\author{
Daniel Shats\inst{1}\orcidID{0000-0001-7861-7834}\and
Hadar Hezi\inst{2}\and
Guy Shani\inst{3}\and
Yosef E. Maruvka\inst{3}\orcidID{0000-0002-8918-9887}\and
Moti Freiman\inst{2}\orcidID{0000-0003-1083-1548}
}
\authorrunning{D. Shats et al.}
%
\institute{Faculty of Computer Science, Technion, Haifa, Israel \and
Faculty of Biomedical Engineering, Technion, Haifa, Israel \and
Faculty of Biotechnology and Food Engineering, Technion, Haifa, Israel \\
\email{shats@campus.technion.ac.il}\\
}
\maketitle

\begin{abstract}
Assessing microsatellite stability status of a patient's colorectal cancer is crucial in personalizing treatment regime. Recently, convolutional-neural-networks (CNN) combined with transfer-learning approaches were proposed to circumvent traditional laboratory testing for determining microsatellite status from hematoxylin and eosin stained biopsy whole slide images (WSI). However, the high resolution of WSI practically prevent direct classification of the entire WSI. Current approaches bypass the WSI high resolution by first classifying small patches extracted from the WSI, and then aggregating patch-level classification logits to deduce the patient-level status. Such approaches limit the capacity to capture important information which resides at the high resolution WSI data.
We introduce an effective approach to leverage WSI high resolution information by momentum contrastive learning of patch embeddings along with training a patient-level classifier on groups of those embeddings. Our approach achieves up to 7.4\% better accuracy  compared to the  straightforward patch-level classification and patient level aggregation approach with a higher stability (AUC, $0.91 \pm 0.01$ vs. $0.85 \pm 0.04$, p-value$<0.01$). Our code can be found at \url{https://github.com/TechnionComputationalMRILab/colorectal_cancer_ai}.

\end{abstract}

\section{Introduction}
Colorectal cancer is a heterogeneous type of cancer which can be generally classified into one of two groups based on the condition of Short Tandem Repeats (STRs) within the DNA of a patient. These two groups are known as Microsatellite Stable (MSS) and Microsatellite Instable (MSI). When the series of nucleotides in the central cores of the STRs exhibit a highly variable number of core repeating units, the patient sample is referred to as Microsatellite Instable. Otherwise, if the number nucleotides is consistent, that patient sample is referred to as Microsatellite Stable (MSS) \cite{li2020microsatellite}. Microsatellite instability (MSI) exists in approximately 15\% of colorectal cancer patients. Due to the fact that patients with MSI have different treatment prospects than those without MSI, it is critical to know if a patient exhibits this pathology before determining treatment direction \cite{boland2010microsatellite}.


Currently, it is possible to determine microsatellite status in a patient by performing various laboratory tests. While these methods are effective, they are expensive and take time, resulting in many patients not being tested for it at all. Therefore, there exist many recent efforts aimed toward utilizing deep-learning-based methods to uncover a computational solution for the detection of MSI/MSS status from hematoxylin and eosin (H\&E) stained biopsy whole slide images (WSI). 

\begin{figure}[ht]
    \centering
    \includegraphics[width=0.75\textwidth]{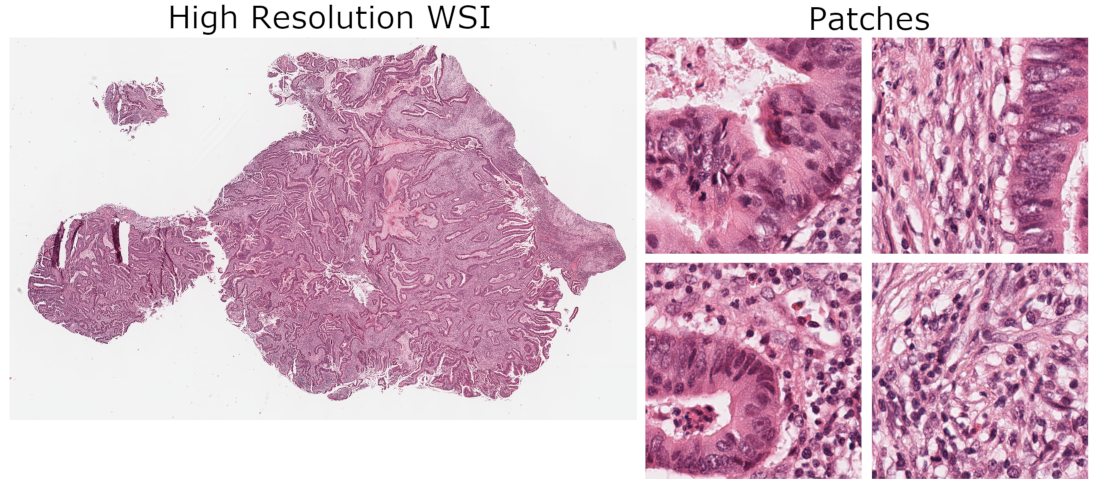}
    \caption{Example of an single patient WSI and a subset of the patches generated from the WSI. The resolution of the WSI is 45815x28384 and the corresponding patches are 512x512.}
    \label{fig:wsi_example}
\end{figure}

However, the WSIs have an extremely large resolution, often reaching over a billion pixels (Fig.~\ref{fig:wsi_example}). Since neural networks can operate, due to computational constraints, only on relatively low resolution data, the input WSI must be shrunk down in some way to a size that is manageable by today's models and hardware. A relatively na\"ive approach to do this is by down sampling inputs to approximately the same resolution as Image Net (which also allows one to leverage transfer learning) \cite{kather2019deep,echle2020clinical}. While this is not detrimental for natural images, where fine detail might not be critical, the same cannot be said for WSI of human tissue where the information of concern may exist in various scales. Our work presents a novel method for effectively down sampling patches into lower dimensional embeddings while preserving features critical to making clinical decisions. Due to our effective dimensional reduction of individual patches, we are then able to concatenate the features of multiple patches together and make clinical decisions using inter-patch information.

\subsection{CNN-based MSI/MSS classification from WSI data}
A simple and straightforward method to alleviate the above issue is by cutting WSI's into many patches and then applying modern deep learning methods to these individual patches as if each of them were a separate image from a data set  \cite{kather2019deep,echle2020clinical}. Once the input has been tessellated, there are a few known ways of training a model from them. One way is to simply assign a label to every patch based on the patient from which they are derived, and then train a model on a patch-by-patch basis. Once the model has been trained, its outputs on all the patches can be averaged for a patient-level classification. Described mathematically, the inference procedure is described below:

Suppose we have some trained classifier $F$ (which returns the probability that a patch belongs to class MSI or MSS e.g. $0\le F(x) \le 1$ for some input patch $x$), a whole slide image $W$, and $n$ patches extracted from $W$ such that $\{p_1, p_2, ... p_n\} \in W$. Then, a patient level probability prediction $P_{W}$ (on the WSI) can be formulated as such:

\begin{equation}\label{eq:1}
    P_{W} = \frac{1}{n}\sum{F(p_n)}
\end{equation}

And given some classification threshold $0 \le t \le 1$, we can arrive at a final classification $C_{W}$ for the patient:

\begin{equation}\label{eq:2}
C_{W} =
    \begin{cases} 
      MSS & P_W < t \\
      MSI & P_W \ge t
    \end{cases}
\end{equation}

However, such approaches practically ignore the fact that much of the information critical to making an informed decision on a patient level may reside in the high resolution and inter-patch space. Further, the classification of the patches based on the patient-level data may result in incorrect classification as not necessarily all patches are contributing equally to the classification of a patient as MSI or MSS.


\subsection{Self-supervision for Patch Embbedings}
In recent years self-supervised learning methods have become an extremely attractive replacement for autoencoders as encoding or downsampling mechanisms, while learning features that are much more informative and meaningful \cite{liu2021self} and therefore can be easily leveraged for use in downstream tasks. This is an exciting property that has led to new research being done in an attempt to circumvent the issues with downsampling described earlier. 

Although self-supervised learning is relatively new and there are many algorithms that attempt to achieve effectively the same goal \cite{grill2020bootstrap,feng2019self,caron2020unsupervised}. Of the various methods, contrastive learning techniques such as those introduced in SimCLR \cite{chen2020simple} have become very popular due to their high efficacy and simplicity.

Recently there have been some attempts to use self-supervised contrastive learning, and more specifically the aforementioned SimCLR algorithm, to aid in classification of WSI imagery. Unfortunately SimCLR requires large computational resources to train in a reasonable amount of time. That is why we decided to test the advent of Momentum Contrast Learning with MoCo v2 by Chen et al \cite{chen2020improved}. This framework relies on storing a queue of previously encoded samples to overcome the large batch size requirement of SimCLR while seemingly improving downstream classification accuracy as well.

\section{Related Work}

\subsection{CNN-based MSS/MSI classification}
Building on top of Echle \cite{echle2020clinical} and the equations \ref{eq:1} and \ref{eq:2} was that of Kather et al \cite{kather2019deep}. Here, transfer learning via pretraining with ImageNet \cite{deng2009imagenet} was employed to marginally improve results of this straightforward method. Although it has an enormous amount of images, they are not medical, and certainly not pictures of H\&E stained biopsy slides. Therefore the degree to which the learned features from ImageNet transfer well to H\&E stained WSI imagery is arguably negligible.

It is also important to discuss the particular resolution under which the patches were acquired. Due to the small size of the patches, any individual patch may not be large enough to contain the information required to make a classification (even on the tissue contained within only that patch). One must understand whether or not the task at hand requires intra-cellular information (requiring maximum slide resolution) or tissue-level information (requiring downsampling before patching). Unfortunately in either case, it is also possible that information at multiple levels of resolution is required for optimal results.

Still, more drawbacks can be found tessellating high resolution images into many smaller patches, regardless of patch resolution concerns. For one, the model cannot learn inter-patch information. This is especially important considering that it is very possible that a majority of the patches do not actually contain targeted tissue. Moreover, training the model in such a way is misleading, considering that many patches which have MSS tissue (yet are found on an MSI classified patient) will be marked as MSI for the model. This is likely to result in the model learning less than optimal features.

The work by Hemati et al \cite{hemati2021cnn} gets around this issue by creating a mosaic of multiple patches per batch during training. Unfortunately, this creates other drawbacks. Most notably, they cannot use all the patches per patient, and so they use another algorithm which is mutually exclusive to the learning procedure in order to choose patches, with no guarantee that they contain targeted tissue. Moreover, the mosaic of these selected patches is also still limited by resolution, and so they still must scale down the patches from their original resolution before training.

An improvement on all these previous works was done in the research by Bilal et al \cite{bilal2021development}. Most notably, they advanced upon the work from Hemati \cite{hemati2021cnn} by learning the patch extraction, or as they call it, patch detection, using a neural network as well. Thereby alleviating an inductive bottleneck. Their process also includes significant work surrounding intermittent detection of known biologically important features to such a problem, such as doing nuclear segmentation, and then providing that information to the next model to make a better-informed decision.

\subsection{Self supervision for patch embbedings}
Due to self-supervised learning being a fairly recent invention, the works similar to ours which cite using it are rather sparse. One of the works which explores the validity of using these methods in the first place is that of Chen et al \cite{chen2022self}. They show that features learned through self-supervised contrastive learning are robust and can be learned from in a downstream fashion. Another paper that uses a similar two stage approach with self-supervised learning being the first stage is DeepSMILE by Schirris et al \cite{schirris2021deepsmile}. They used a similar approach to the above mentioned contrastive self-supervised learning step with SimCLR, but learned on the features using Deep Multiple Instance Learning (MIL) \cite{ilse2018attention}. While this approach was effective, the computational requirements of SimCLR and the added complexity of MIL may keep the advent of this research out of the hands of many researchers.

Very recently an improvement on the work by Chen \cite{chen2022self} was introduced in their research using Hierarchical Vision transformers \cite{chen2022scaling}. Here, the authors apply self-supervised learning through DINO \cite{caron2021emerging} to train 3 stages of vision transformers in achieving entire WSI level classifications. Though seemingly effective, the increased complexity of their approach and the necessity of utilizing transformers makes it relatively inflexible.

\subsection{Hypothesis and Contributions}
\paragraph{Hypothesis:} Learning effective patch embeddings with self-supervised learning and training a small classifier on groups of those embeddings is more effective than either training on down-sampled WSI's or training on individual patch embeddings and averaging the classification for a patient.

We believe this hypothesis to be true due to the ability of a network to learn inter-patch information at an embedding level. This way, information that is encoded in one patch can impact the decision of the entire WSI. Our contribution is an intuitive and elegant framework that improves patient classification accuracy by up to 7.4\%. We argue that this method is very simple to understand and has many avenues of possible improvement. Specifically when considering the initial feature extraction stage, there are many other self-supervised representation learning methods that can be tested and directly compared using our approach.

\section{Data}
The training and validation data used in our method consists of the COAD and READ cohorts of The Cancer Genome Atlas (TCGA) \cite{weinstein2013cancer}. Out of a total of 360 unique patients, the train set is comprised of 260 patients and the validation set is comprised of 100 patients, where each patient is equivalent to one WSI. Each of these WSIs were tessellated into patches of size 512x512 pixels at  0.5 $\mu$m/px and then downsampled to 224x224 (this was done only for comparison with Kather et al \cite{kather2019deep}). Next the probability of each patch to contain cancerous tissue was computed by a trained CNN and only the patches which were likely to have cancer tissue were kept. Finally, the patches were color normalized. Further detail on the data preprocessing procedure can be found in the paper by Kather et al \cite{kather2019histological}. The 260 train patients were tessellated into 46,704 MSS labeled patches, and 46,704 MSIMUT labeled patches. The 100 validation patients were tessellated into 70,569 MSS labeled patches, and 28,335 MSIMUT labeled patches.

Finally, we also ran some final experiments on a more balanced subset of the validation dataset, referred to later in this paper as the “Balanced Validation Set”. It comprises of 15 MSS patients and 15 MSIMUT patients which all have a relatively similar distribution of patches extracted from them. 7281 patches were extracted from the MSIMUT patients and 7446 patches were extracted from the MSS patients.

\section{Method and Model}

\subsection{Overview}
Our method comprises of two main training stages:

\begin{figure}[H]
    \centering
    \includegraphics[width=0.99\textwidth]{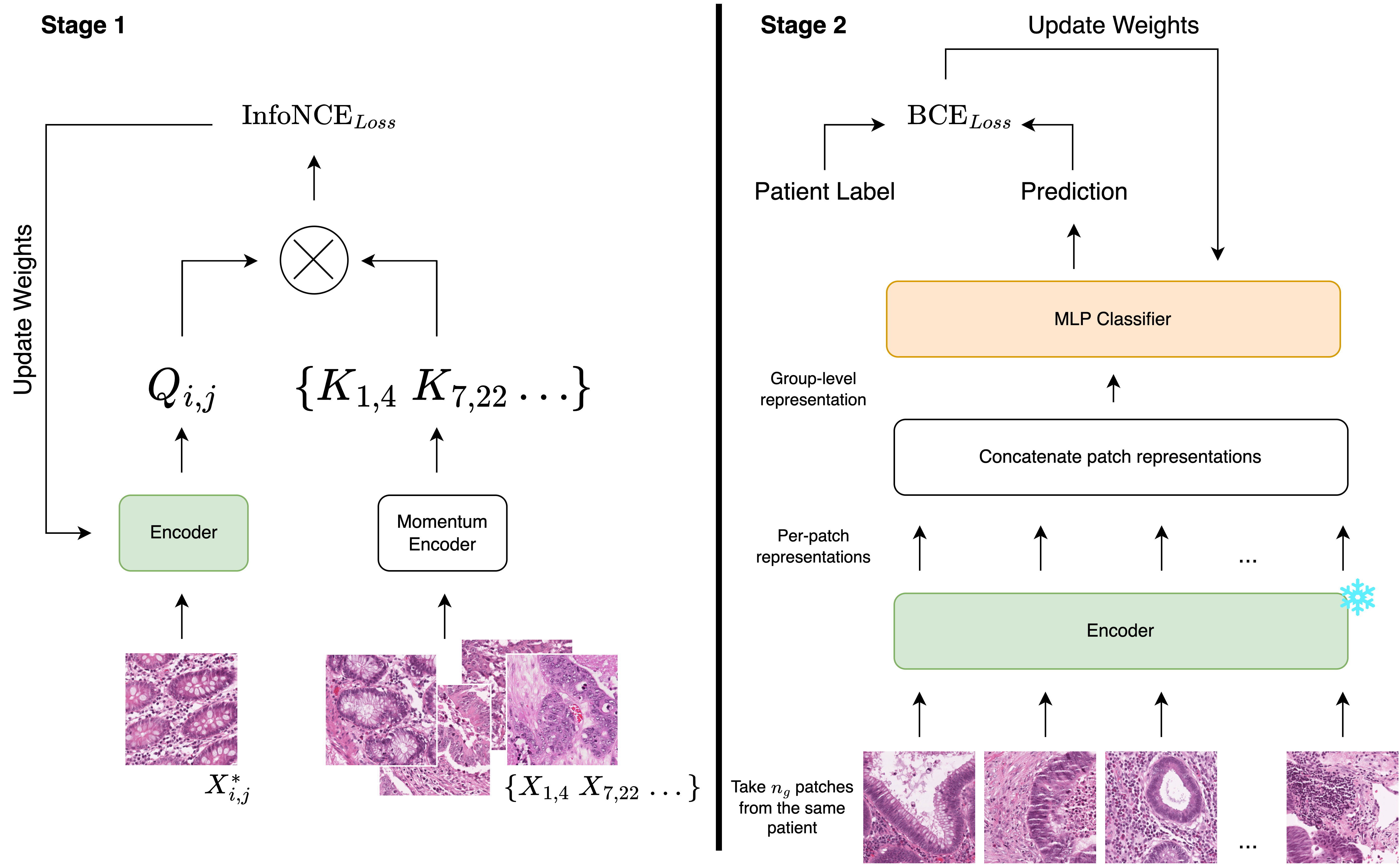}
    \caption{Both stages of our proposed model.}
    \label{fig:learning_diagram}
\end{figure}

Stage 1 utilizes MoCo and stands to generate robust patch level embeddings that encode patches in a way such that they can be learned from in a downstream fashion. Above, $Q_{i,j}$ represents the encoded query for patient $i$ and patch $j$. Similarly $K_{i, j}$ represents the same for patches encoded by the momentum encoder stored in the queue, and $X_{i,j}$ corresponds to individual samples from the dataset relating to patient $i$ and patch $j$. The stage 1 diagram is very similar to that found in Chen et al \cite{chen2020improved}.

Stage 2 groups the patches so that their features can be aggregated and the head of our model can learn from a set of patches as opposed to an individual sample. The encoder in stage 2 is a frozen copy of the trained encoder from stage 1. The snowflake indicates that its gradients are not tracked.

\subsection{Stage 1: Training a self-supervised Feature Extractor for Patch-level Embeddings}
In stage 1 of training, our feature extractor is trained in exactly the same way as described in the MoCo v2 paper \cite{chen2020improved}. Data loading and augmentation are unchanged. The main difference is our use of a Resnet18 \cite{he2016deep} backbone as opposed to the Resnet50 (C4) backbone which was tested in the original implementation. This was done due to computational constraints and for comparison to the baseline approach from Kather. We also used cosine learning rate annealing, which seems to improve training. The output dimension of our feature extractor is 512 ($n_o$).

To evaluate the ability of MoCo to extract usable features from patches, we tracked the value of the InfoNCE loss \cite{oord2018representation} on the training set. After achieving the lowest value for train InfoNCE loss (0.88 in our case), the model was saved and used for stage 2 training. This was not tracked on the validation dataset to avoid overfitting. Below you can see the training curve for MoCo over 621 epochs:

\begin{figure}[h]
    \centering
    \includegraphics[width=0.65\textwidth]{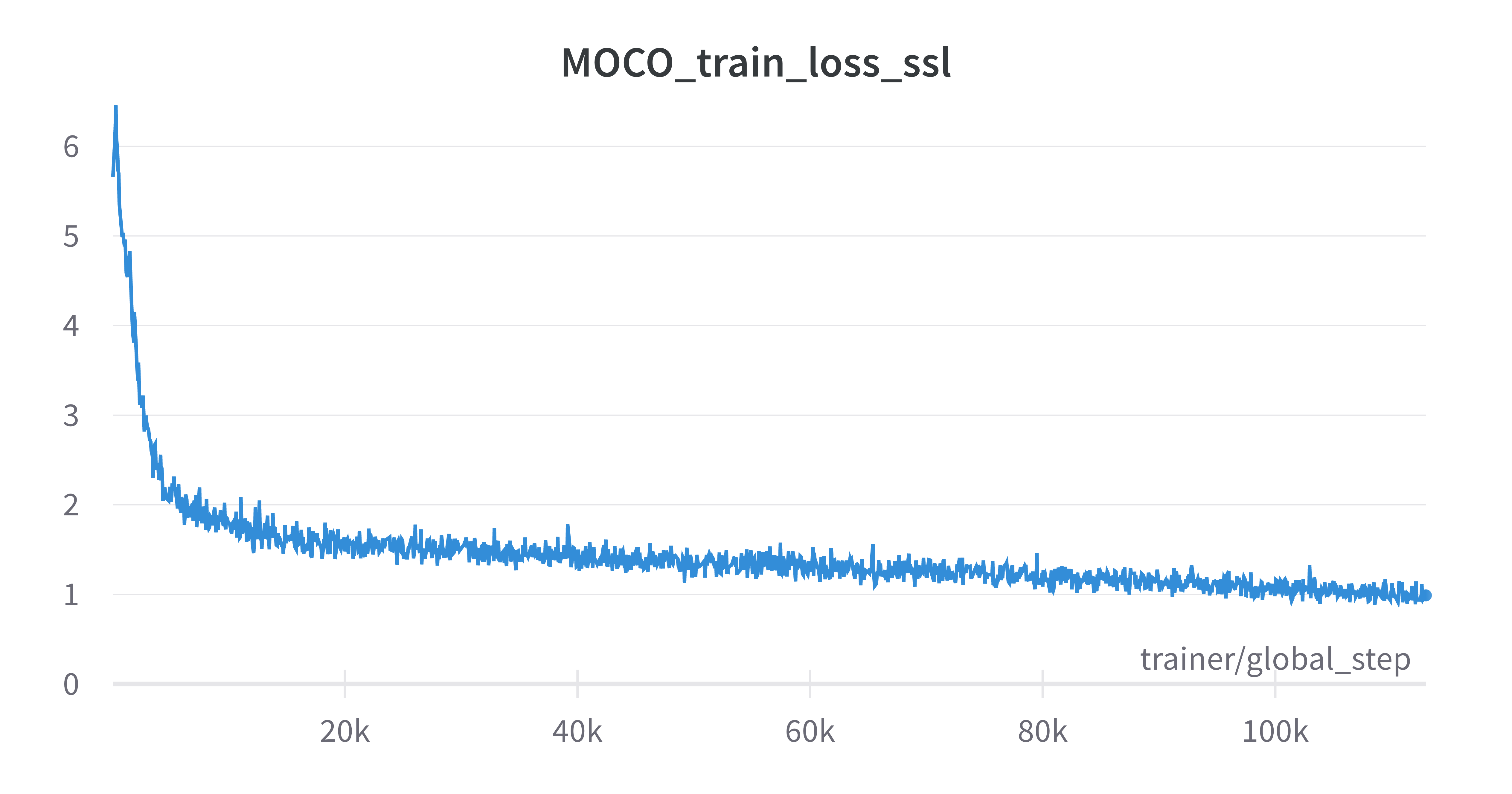}
    \caption{InfoNCE training loss over 621 epochs for MoCo.}
    \label{fig:infonce_loss}
\end{figure}

The goal of contrastive learning, and thus momentum contrast learning, is to (as stated in the paper by He et al) learn an encoder for a dictionary look up task. The contrastive loss function InfoNCE exists to return a low loss when encoded queries to the dictionary are similar to keys, and a high loss otherwise.

\subsection{Stage 2: Training a Supervised Classifier on Patch Embedding Groups}
In stage 2, the resnet18 \cite{he2016deep} feature extractor trained by MOCO is frozen, and so gradients are not tracked. When making the forward pass, features extracted from patches are grouped by $n_g$ (group size), meaning they are concatenated into one long vector. The length of this vector ($l_g$) will be:

\begin{equation}\label{eq:final_group_vec_len}
    l_{g} = n_{g}*n_{o}
\end{equation}

Meaning that the input dimension of our multi-layer percpetron (MLP) group-level classifier, or model head that we are training in stage 2 must have an input dimension of $l_g$. This brings us to the first issue regarding the $n_g$ parameter. The larger the group size after feature extraction, the larger the first layer of the head must be. This is likely why we found a group size of 4 to be optimal for our dataset. When using a larger group size, the number of parameters for the head of the model increases dramatically, and it tends to overfit much faster.

As an interesting test, we also attempted $n_g = 1$, which performed very similarly to the standard approach. This is what we expected as it indicates the embedding space from Momentum Contrastive Learning is similarly effective to the embedding space of a model trained in a supervised fashion.

\subsection{Evaluation}
Judgement of the algorithm is performed using two main criteria. The first is patch level accuracy and the second is patient level accuracy. 

Patch level accuracy ($A_{patch}$) is exactly the same as accuracy in the general context. The only caveat is how the patches are assigned their label. Since our WSI's are labeled on a patient basis, the patches are labeled by inheriting the label of the patient to which they belong.

\begin{equation}
    A_{patch} = \frac{\textrm{Number of Correctly Predicted Patches}}{\textrm{Total Number of Patches}}
\end{equation}

Patient level accuracy ($A_{patient}$) is a more crucial and more difficult to improve upon metric. It cannot be trained for directly, as an entire WSI cannot fit on GPU without downsampling. To measure this metric, we must save the models predictions on individual patches (or on groups of patches and extrapolate individual patch predictions) and calculate a final prediction for a patient using the cumulative predictions of its constituent patches. This can be done using a majority vote approach or it is also possible to treat each patches prediction as a probability and average the probabilities before thresholding on a patient level and achieving a final prediction.

\begin{equation}
    A_{patient} = \frac{\textrm{Number of Correctly Predicted Patients}}{\textrm{Total Number of Patients}}
\end{equation}

\section{Experiments and Results}

\subsection{Standard Dataset}

The results in this section refer to performance measured on the original dataset processed by Kather et al \cite{kather2019histological}. We show the results of our implementation of Kather's method compared to our improvement using momentum contrastive learning and group patch embeddings.

\setlength{\tabcolsep}{4pt}
\begin{table}[H]%
    \centering
    \caption[Data description]{Accuracy Comparison on standard dataset. Both methods trained to 100 epochs. These are the validation results of an average of 10 runs per method. Our method achieves significantly higher accuracy in both patient (paired t-test, p$<0.001$), and patch level (paired t-test, p$<<1e-6$) evaluation while also having a much more stable result given its smaller standard deviation. }
    \begin{tabular}{l|lc}
        \hline
        \noalign{\smallskip}
        Method      &Patient Accuracy & Patch Accuracy\\
        \noalign{\smallskip}
        \hline
        Ours            & $\textbf{0.862}\pm0.006$   &  $\textbf{0.797}\pm0.005$\\
        Kather et al.   & $0.837\pm0.016$   & $0.716\pm0.015$ \\ \hline
    \end{tabular}
    \label{tab:accuracy_standard_dataset}
\end{table}
\setlength{\tabcolsep}{1.4pt}

\begin{figure}[H]
     \centering
     \begin{subfigure}[b]{0.49\textwidth}
         \centering
         \includegraphics[width=\textwidth]{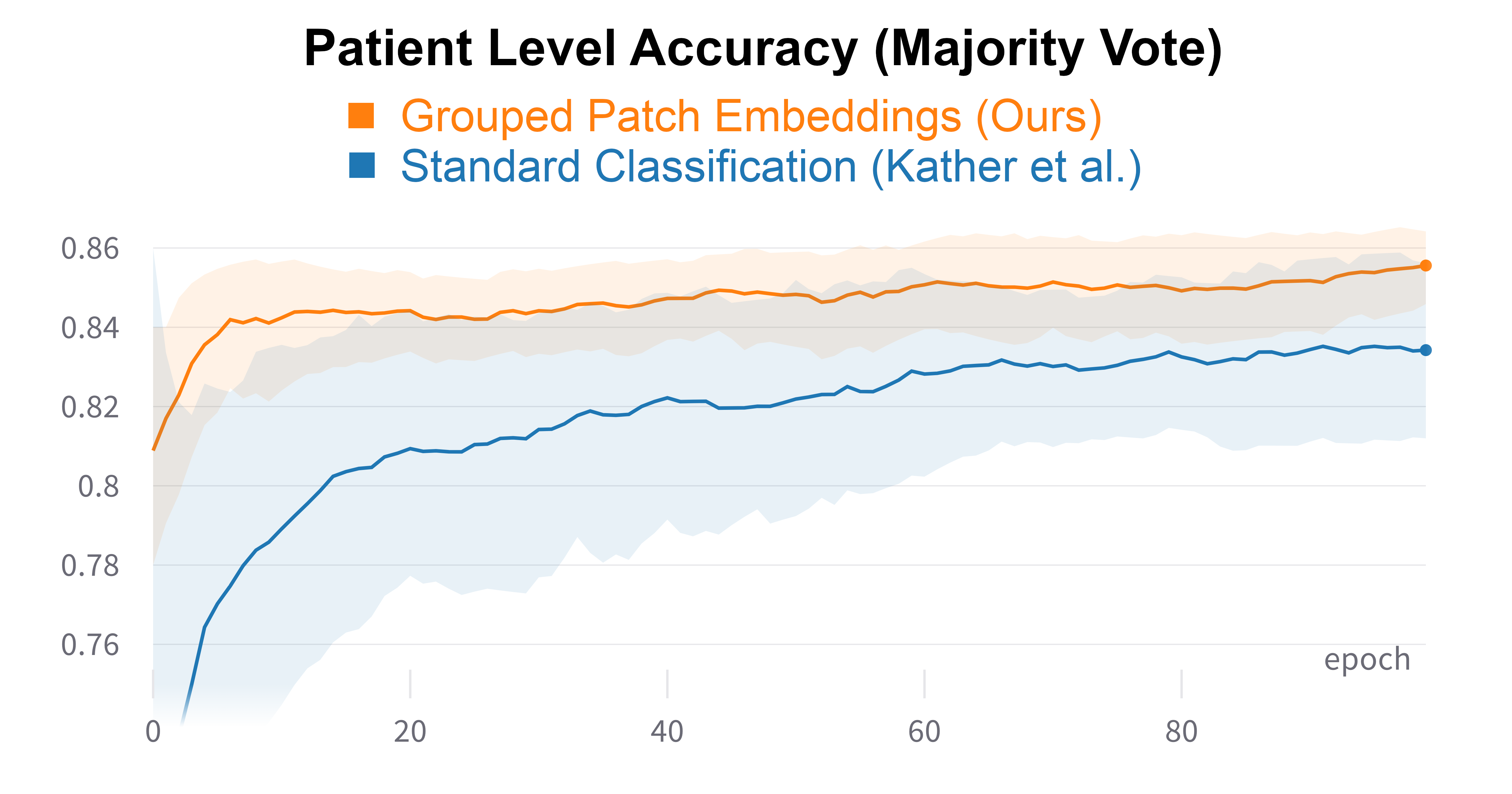}
         \label{fig:patient_accuracy_training_curve}
     \end{subfigure}
     \begin{subfigure}[b]{0.49\textwidth}
         \centering
         \includegraphics[width=\textwidth]{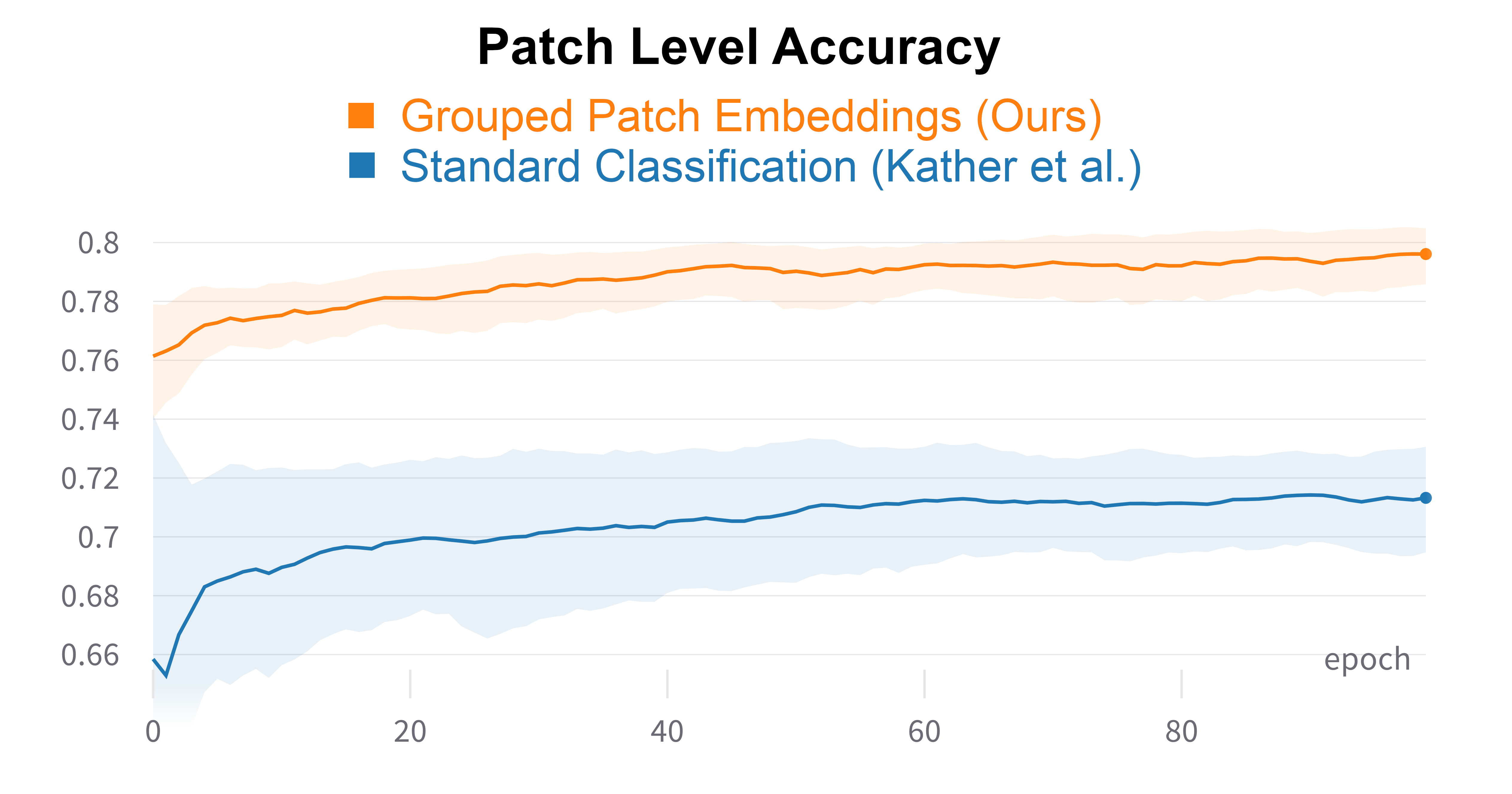}
         \label{fig:patch_accuracy_training_curve}
     \end{subfigure}
        \caption{Validation Accuracy during training. Average of 10 runs.}
        \label{fig:accuracy_training_curves}
\end{figure}
Due to the feature extractor already having been learned, our method initially trains much faster than the baseline. We have even noted that for some hyperparameter combinations it may be most effective to stop training after only a few epochs. And below are the ROC curves for the above models:
\begin{figure}[H]
     \centering
     \begin{subfigure}[b]{0.4\textwidth}
         \centering
         \includegraphics[width=\textwidth]{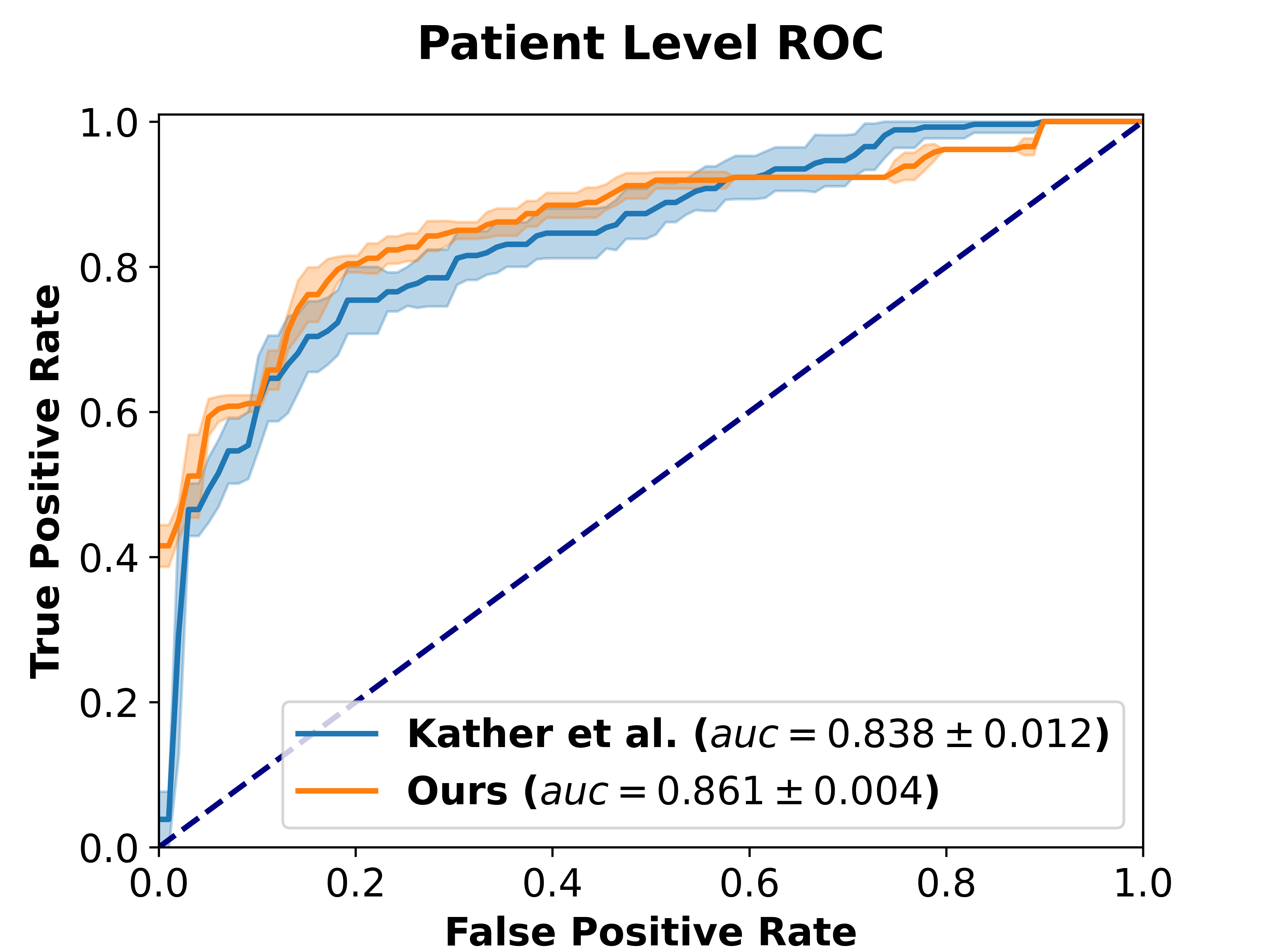}
         \label{fig:rocs_standard_dataset_patient_level}
     \end{subfigure}
     \begin{subfigure}[b]{0.4\textwidth}
         \centering
         \includegraphics[width=\textwidth]{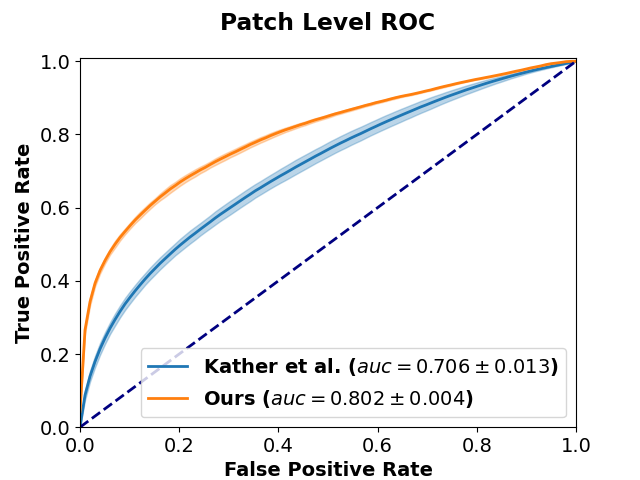}
         \label{fig:rocs_standard_dataset_patch_level}
     \end{subfigure}
        \caption{ROC curves on standard validation dataset. Average of 10 runs. Our method achieves a significantly higher AUC in both patient (paired t-test, p$<0.01$) and patch level evaluation (paired t-test, p$<<1e-7$).}
        \label{fig:rocs_standard_dataset}
\end{figure}

\subsection{Balanced Validation Set}
The results in this section refer to performance measured on the balanced subset of the validation dataset processed by Kather et al \cite{kather2019histological}. We describe the composition of this dataset in the second paragraph of the data section of this paper. Our method does even better on this balanced validation set compared to the original one from above. The differences were significant for both patient level (paired t-test, p$<<1e-4$), and patch level (paired t-test, p$<<1e-6$) classification. This suggest that our method is less prone to bias and overfitting to patients with more or less patches that have been extracted from them.

\setlength{\tabcolsep}{4pt}
\begin{table}[ht]%
    \centering
    \caption[Data description]{Accuracy Comparison on balanced dataset.}
    \begin{tabular}{l|lc}
        \hline
        \noalign{\smallskip}
        Method      &Patient Accuracy & Patch Accuracy\\
        \noalign{\smallskip}
        \hline
        Ours            & $\textbf{0.797}\pm0.010$   &  $\textbf{0.751}\pm0.006$\\
        Kather et al.   & $0.723\pm0.026$   & $0.662\pm0.013$ \\ \hline
    \end{tabular}
    \label{tab:accuracy_balanced_dataset}
\end{table}
\setlength{\tabcolsep}{1.4pt}

\begin{figure}[H]
     \centering
     \begin{subfigure}[b]{0.4\textwidth}
         \centering
         \includegraphics[width=\textwidth]{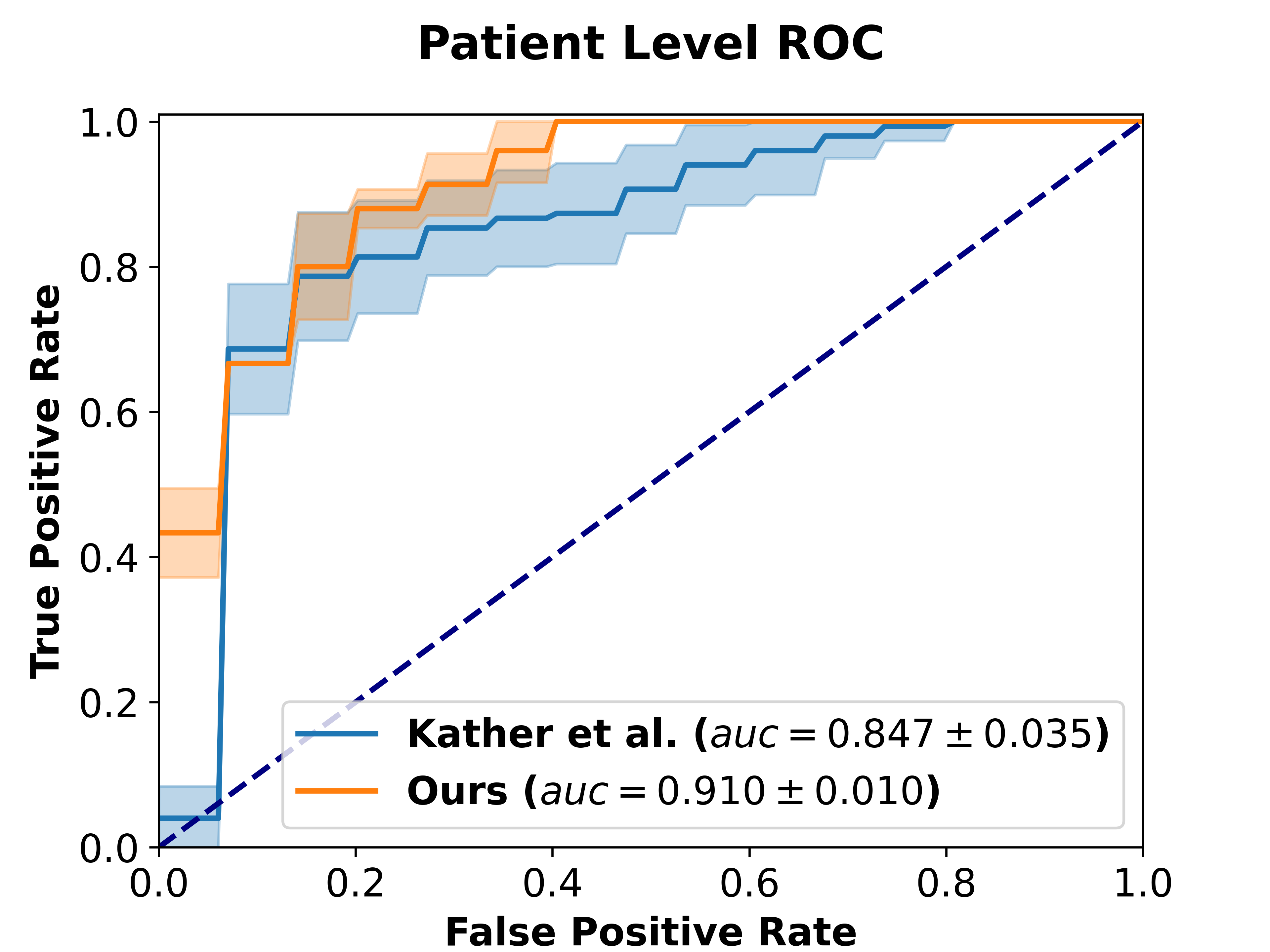}
         \label{fig:rocs_balanced_dataset_patient_level}
     \end{subfigure}
     \begin{subfigure}[b]{0.4\textwidth}
         \centering
         \includegraphics[width=\textwidth]{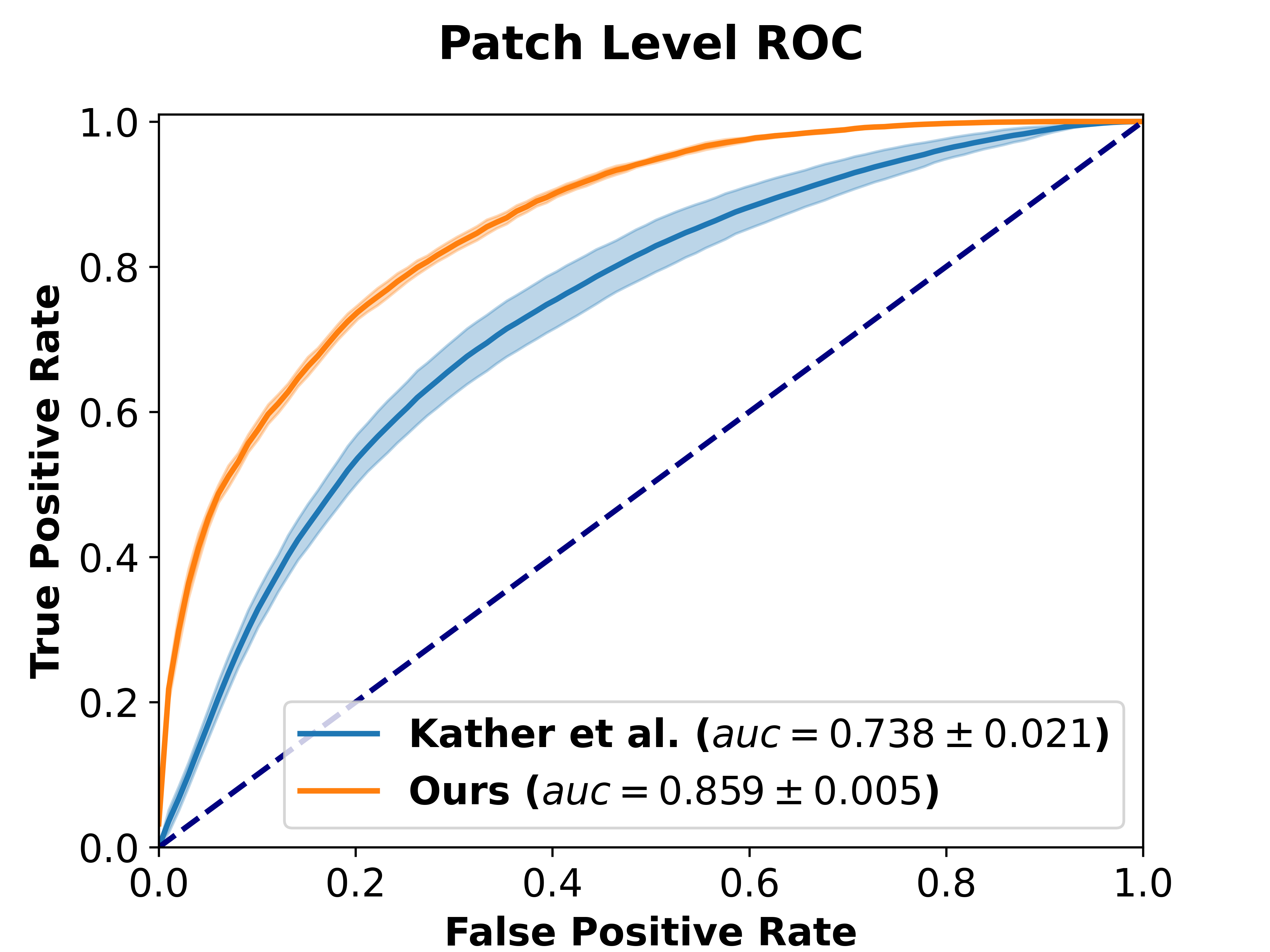}
         \label{fig:rocs_balanced_dataset_patch_level}
     \end{subfigure}
        \caption{ROC curves on balanced dataset. Average of 10 runs. Our method achieves a significantly higher AUC in both patient (paired t-test, p$<0.01$) and patch level evaluation (paired t-test, p$<<1e-6$).}
        \label{fig:rocs_balanced_dataset}
\end{figure}
\section{Conclusions}

Our work validates the feasibility of learning usable features from H\&E stained biopsy slide patches using momentum contrast learning. We also qualify that learning from the aggregated features of multiple patches works better than simply averaging the predictions of individual patches for a WSI prediction. Finally, we contribute a simple and intuitive framework for combining these concepts with huge potential for improvement. The future for this domain lies in improving patch level feature extraction and aggregating more features to make global WSI decisions. The advent of a WSI classifier that is as accurate as laboratory testing for microsatellite status can drastically improve the rate at which patients are diagnosed and their treatment prospects.

%
%
\bibliographystyle{splncs04}
\bibliography{egbib}
\end{document}